\documentclass[11pt]{article}

\usepackage[utf8]{inputenc}
\usepackage[T1]{fontenc}
\usepackage{lmodern}
\usepackage{microtype}
\usepackage[a4paper,margin=2.5cm]{geometry}
\usepackage{amsmath,amssymb,amsthm,mathtools}
\usepackage{bm}
\usepackage{graphicx}
\usepackage{booktabs}
\usepackage{multirow}
\usepackage{array}
\usepackage{xcolor}
\usepackage[ruled,vlined,linesnumbered]{algorithm2e}
\usepackage[numbers,sort&compress]{natbib}
\usepackage[colorlinks=true,linkcolor=blue,citecolor=blue,urlcolor=blue]{hyperref}
\newcommand{\orcidlink}[1]{\href{https://orcid.org/#1}{\raisebox{-0.15ex}{\includegraphics[height=1.9ex]{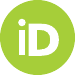}}}}
\usepackage{cleveref}
\usepackage{placeins}

\graphicspath{{figures/}}


\newtheorem{theorem}{Theorem}
\newtheorem{proposition}[theorem]{Proposition}

\theoremstyle{definition}

\newtheorem{assumption}{Assumption}
\theoremstyle{remark}

\newcommand{\R}{\mathbb{R}}
\newcommand{\E}{\mathbb{E}}
\newcommand{\Pb}{\mathbb{P}}
\newcommand{\D}{\mathcal{D}}
\newcommand{\ind}{\mathbf{1}}
\newcommand{\eaml}{\textup{\textsc{EAML}}}
\DeclareMathOperator*{\argmax}{arg\,max}

\newcommand{\doi}[1]{\href{https://doi.org/#1}{doi:\,#1}}

\title{Evolutionary Adaptive Metric Learning: A Multi-Objective Framework for Learning Local Distance Functions in Nearest Neighbor Classification}

\author{%
  Isaac Touza\textsuperscript{1,2,$*$}\,\orcidlink{0009-0003-6288-9458}\quad
  \\[2ex]
  \parbox{0.92\linewidth}{\centering\small
    \textsuperscript{1}Department of Mathematics and Computer Science, Faculty of Science, University of Maroua, P.O.~Box 814, Maroua, Cameroon\\[0.4ex]
    \textsuperscript{2}Laboratoire de Recherche en Informatique (LaRI), University of Maroua, Maroua, Cameroon\\[0.4ex]
    }%
}
\date{}

\begin{document}
\maketitle
{\renewcommand{\thefootnote}{\fnsymbol{footnote}}%
\footnotetext[1]{Corresponding author. E-mail: \href{mailto:isaac_touza@outlook.fr}{isaac\_touza@outlook.fr}.}}

\begin{abstract}
Nearest neighbour classifiers inherit the geometry of the distance they use.
Fixed distances (Euclidean, Manhattan, Mahalanobis) and global learned
metrics (NCA, LMNN, ITML) apply the same geometry everywhere, whereas the
relevant directions of real data may change from one region of the input
space to another. We propose \eaml{} (Evolutionary Adaptive Metric
Learning), which attaches to every point $x$ a positive-definite matrix
$M_\theta(x)=D_\theta(x)+UU^\top+\epsilon I$. Here $D_\theta(x)$ is a diagonal
scaling interpolated between a few anchors by a softmax gating, and $U$ is a
global low-rank correlation term. The matrices are not learned one by one: a
real-coded elitist genetic algorithm evolves the low-dimensional parameters
that generate them. It minimises a continuous scalarisation of four
objectives: a leave-one-out stochastic-neighbour error, a robustness hinge on
relative hit/miss margins, the smoothness of the metric field, and a shrinkage
penalty. We prove that the learned matrices are uniformly positive definite and
equivalent to the Euclidean metric, that an optimal metric exists, and that
the metric field is Lipschitz. We also prove that $k$-NN decisions are stable
under perturbations smaller than an explicit margin, give a uniform
generalisation bound whose complexity term depends on the number of evolved
parameters rather than on the sample size, and show that the elitist GA
converges in value to the global minimum almost surely. Empirically, on a
controlled problem whose relevant feature switches between two regions, \eaml{}
recovers the region-dependent weights and is the most accurate method (93.2\%
vs.\ 91.4\% for the best global metric). On thirteen public datasets under a
nested $5\times5$-fold protocol, the picture is more nuanced: \eaml{}
significantly improves on the Euclidean metric on Glass ($+7.0$ points) and
Ionosphere ($+4.6$), but its average rank is not better than that of the global
baselines. No method differs significantly from the others overall (Friedman
$p=0.14$). Under nuisance features or test-time noise, a global evolved diagonal
metric is preferable. We report these negative results together with
diagnostics (unfinished search, gating misalignment, sharpened metrics with
$k=1$) that delimit when local metrics pay off.
\end{abstract}


\section{Introduction}

The $k$-nearest neighbour ($k$-NN) rule \citep{cover1967} classifies a point
by looking at its closest training examples, so its behaviour is entirely
determined by the notion of closeness. The Euclidean distance treats all
features alike. The Mahalanobis distance \citep{mahalanobis1936} rescales and
decorrelates them using the global covariance. Metric learning goes further
and fits a Mahalanobis matrix to the labels: from side information
\citep{xing2002}, by maximising a stochastic leave-one-out score (NCA,
\citealp{goldberger2004}), with large-margin constraints (LMNN,
\citealp{weinberger2009}) or with an information-theoretic regulariser (ITML,
\citealp{davis2007}); see \citet{kulis2013,bellet2013survey} for surveys.

All these metrics are \emph{global}. They apply a single linear geometry to
the entire input space, yet in real data the relevant features, their
correlations and the noise level can depend on the region. A feature that
separates two classes in one part of the space may be pure noise elsewhere.
Locally adaptive rules address this. DANN \citep{hastie1996} and the method
of \citet{domeniconi2002} reshape the neighbourhood of each query from local
statistics. Multi-metric LMNN \citep{weinberger2009} learns one metric per
class or cluster. Parametric local metric learning (PLML, \citealp{wang2012})
combines basis metrics anchored in the input space with manifold
regularisation. Adaptive neighbourhood sizes form a related line of work
\citep{mullick2018,levada2024}.

\paragraph{Question.} Can we learn, \emph{jointly}, a local metric field, its
smoothness and its robustness, while keeping the complexity of the learned
object under explicit mathematical control? And does local adaptivity actually
help on standard benchmarks once every competitor is tuned with the same
care? We study these questions with an evolutionary approach. The
optimisation problem combines non-convex, non-differentiable-looking terms
(margins, minima over training points) with box constraints, and a GA handles
it without gradients.

\paragraph{Contributions.}
\begin{enumerate}
\item \textbf{Model.} We introduce a family of local metrics
  $M_\theta(x)=D_\theta(x)+UU^\top+\epsilon I$ generated by $D=Rp+pq$ parameters,
  with $R$ anchors and rank $q$, independent of the sample size $n$. The family
  contains the Euclidean metric, global diagonal metrics ($R=1$) and global
  low-rank corrections (\Cref{sec:method}).
\item \textbf{Algorithm.} \eaml{} evolves these parameters with an elitist
  real-coded GA \citep{deb1995,deb2001}. Its fitness is a continuous
  scalarisation of four objectives: classification, robustness, smoothness and
  complexity (\Cref{sec:method}).
\item \textbf{Theory.} We prove positivity and metric equivalence, existence of
  an optimal metric, Lipschitz regularity of the metric field, robustness of
  $k$-NN decisions to perturbations, a uniform generalisation bound and the
  almost-sure convergence in value of the GA (\Cref{sec:theory}).
\item \textbf{Evidence.} We compare \eaml{} with seven baselines (Euclidean,
  Manhattan, Mahalanobis, NCA, LMNN, ITML and an evolved global diagonal metric)
  on a controlled problem and on thirteen datasets, including a Fashion-MNIST
  subsample \citep{xiao2017}, with nested $5\times5$-fold cross-validation,
  noise-robustness tests and ablations (\Cref{sec:experiments}). We report where
  the local model helps and where it does not.
\end{enumerate}

\section{Related work}\label{sec:related}

\paragraph{Global metric learning.} Most supervised metric learning methods
learn a single positive semi-definite matrix $M$ such that
$d_M(x,z)^2=(x-z)^\top M(x-z)$ brings same-class points together. Xing et
al.\ \citep{xing2002} posed it as a convex problem with pairwise constraints.
NCA \citep{goldberger2004} maximises the expected LOO accuracy of a stochastic
neighbour rule, which is the classification term we also use. LMNN
\citep{weinberger2009} enforces large margins between target neighbours and
impostors. ITML \citep{davis2007} regularises $M$ towards a prior with a LogDet
divergence. Generalisation guarantees for metric learning have been derived
through algorithmic robustness \citep{bellet2015}.

\paragraph{Local metrics and adaptive neighbourhoods.} DANN
\citep{hastie1996} and ADAMENN \citep{domeniconi2002} estimate a query-specific
metric from local class statistics. Multi-metric LMNN \citep{weinberger2009}
assigns one metric per cluster. PLML \citep{wang2012} expresses each local metric
as a combination of anchored basis metrics with smooth weights, and our gating
construction is close in spirit. Closely related are methods that adapt the
neighbourhood \emph{size} rather than its shape, for instance with neural
networks \citep{mullick2018} or with curvature estimates of the data manifold
\citep{levada2024}.

\paragraph{Evolutionary metric and feature weighting.} GAs were used early
on to learn feature weights for $k$-NN \citep{kelly1991}, and later for joint
feature selection, weighting and extraction \citep{raymer2000}. These approaches
evolve a \emph{global} diagonal metric, which corresponds to our
\textsc{GA-global} baseline ($R=1$, $q=0$). Multi-objective evolutionary
algorithms such as NSGA-II \citep{deb2002} return Pareto fronts. We use a
scalarisation instead and leave the Pareto analysis for future work. Our GA
analysis follows the classical argument for elitist algorithms
\citep{rudolph1994}.

\section{The EAML model and algorithm}\label{sec:method}

\subsection{A parametric family of local metrics}

Let $\D=\{(x_i,y_i)\}_{i=1}^n\subset\R^p\times\{1,\dots,C\}$, with standardised
features. We choose $R$ anchors $c_1,\dots,c_R$ (the $k$-means centres of the
training fold) and the gating
\begin{equation}
\pi_r(x)=\frac{\exp(-\|x-c_r\|^2/\tau)}{\sum_{s=1}^R\exp(-\|x-c_s\|^2/\tau)},
\end{equation}
where $\tau$ is the median squared distance from a training point to its
closest anchor. Each anchor carries a vector of log-scales $s_r\in[-B,B]^p$.
The local log-scale field and the local metric are
\begin{equation}
h_\theta(x)=\sum_{r=1}^R\pi_r(x)\,s_r,\qquad
M_\theta(x)=\underbrace{\operatorname{diag}\big(e^{h_\theta(x)}\big)}_{D_\theta(x)}+UU^\top+\epsilon I,
\label{eq:metric}
\end{equation}
with $U\in[-B_U,B_U]^{p\times q}$. This is the factorisation
$M_\theta=L_\theta L_\theta^\top+\epsilon I$ with
$L_\theta(x)=[D_\theta(x)^{1/2},U]$. The distance from a query $x$ to a reference
point $z$ uses the metric of the query:
\begin{equation}
d_\theta^2(x,z)=(x-z)^\top M_\theta(x)(x-z).
\end{equation}
The genome is $\theta=(s_1,\dots,s_R,U)$, of dimension $D=Rp+pq$. The case
$R=1,q=0$ is a global diagonal metric, and $\theta=0$ gives the Euclidean metric
(up to the factor $1+\epsilon$).

\subsection{Objective}

Let $S\subseteq\{1,\dots,n\}$ be the query points used in the fitness: all
points, or a random subset of size $s=800$ when $n>800$. We minimise
\begin{equation}
\mathcal J(\theta)=\mathcal L_{\mathrm{cls}}+\lambda_1\mathcal L_{\mathrm{rob}}+\lambda_2\mathcal L_{\mathrm{sm}}+\lambda_3\,\mathcal C,
\label{eq:objective}
\end{equation}
with the following four terms.
\begin{itemize}
\item \textbf{Classification} (stochastic-neighbour LOO error, as in NCA):
  \[
  \mathcal L_{\mathrm{cls}}=1-\frac1{|S|}\sum_{i\in S}\sum_{j\ne i:\,y_j=y_i}p_{ij},\qquad
  p_{ij}=\frac{\exp(-d^2_\theta(x_i,x_j))}{\sum_{l\ne i}\exp(-d^2_\theta(x_i,x_l))}.
  \]
\item \textbf{Robustness} (relative margin hinge): with $d^{\mathrm h}_i$ and
  $d^{\mathrm m}_i$ the distances from $x_i$ to its nearest same-class and
  other-class training points, and $\rho_i=(d^{\mathrm m}_i-d^{\mathrm h}_i)/(d^{\mathrm m}_i+d^{\mathrm h}_i)$,
  we set $\mathcal L_{\mathrm{rob}}=\frac1{|S|}\sum_i\max(0,\bar\rho-\rho_i)/\bar\rho$
  with target $\bar\rho=0.2$. \Cref{thm:robust} links these margins to
  decision stability.
\item \textbf{Smoothness} of the metric field on the Euclidean $5$-NN graph of $S$:
  \[
  \mathcal L_{\mathrm{sm}}=\frac{1}{5|S|p(2B)^2}\sum_{i\in S}\sum_{j\in N_5(i)}\|h_\theta(x_i)-h_\theta(x_j)\|^2 .
  \]
\item \textbf{Complexity} (shrinkage towards the Euclidean metric):
  $\mathcal C=\frac{1}{Rp\,B^2}\sum_r\|s_r\|^2+\frac{1}{pq\,B_U^2}\|U\|_F^2$.
\end{itemize}
We use $\lambda_1=0.5$, $\lambda_2=0.1$ and $\lambda_3=0.01$ on every dataset. All
terms are continuous in $\theta$, which is what \Cref{thm:existence,thm:ga}
require.

\subsection{Genetic search}

\Cref{alg:eaml} summarises the method. The GA uses a population of 30 genomes
and 40 generations, i.e.\ at most $1230$ objective evaluations. Selection is by
tournament of size 3. Simulated binary crossover \citep{deb1995,deb2001} is
applied with probability $0.9$ and index $15$. Mutation is Gaussian with
per-gene probability $0.1$ (at least one gene per offspring) and a step that
anneals from $0.25$ to $0.02$ of the box width. Offspring are reflected into
the box $\Theta$. The two best parents survive, and the run stops early after
15 generations without improvement. The initial population contains the
Euclidean genome $\theta=0$ and 29 genomes drawn uniformly in $[-0.05,0.05]$
times the box. After evolution, the number of neighbours $k\in\{1,\dots,25\}$ is
chosen by LOO accuracy of the $k$-NN rule under the learned metric, exactly as
for all baselines.

\begin{algorithm}[t]
\caption{\eaml{}: Evolutionary Adaptive Metric Learning}\label{alg:eaml}
\KwIn{standardised training set $\D$; anchors $R$; rank $q$; weights $\lambda_{1:3}$; GA settings}
\KwOut{metric parameters $\hat\theta$ and number of neighbours $\hat k$}
Compute $R$ anchors by $k$-means, the temperature $\tau$ and the gating $\pi(x_i)$\;
Select the query set $S$ and the Euclidean $5$-NN graph of $S$\;
Initialise the population with $\theta=0$ (Euclidean) and small random genomes\;
\For{$g=1,\dots,G$}{
  Tournament selection; SBX crossover; annealed Gaussian mutation; reflection into $\Theta$\;
  Evaluate $\mathcal J(\theta)$ for every offspring (two matrix products of size $|S|\times p\times n$)\;
  Elitist replacement\;
  \lIf{no improvement during 15 generations}{\textbf{break}}
}
$\hat\theta\gets$ best genome; $\hat k\gets\argmax_k$ LOO accuracy of $k$-NN under $d_{\hat\theta}$\;
\BlankLine
\textbf{Prediction} for $x$: compute $\pi(x)$, $M_{\hat\theta}(x)$, the distances $d_{\hat\theta}(x,x_j)$, and return the majority label of the $\hat k$ nearest neighbours.
\end{algorithm}

\section{Theoretical analysis}\label{sec:theory}

Throughout this section the training sample, the anchors $c_r$ and the
gating temperature $\tau$ are fixed. We assume the following.

\begin{assumption}\label{ass:domain}
The (standardised) inputs lie in the ball $\mathcal X=\{x:\|x\|_2\le\varrho\}$,
the anchors satisfy $\|c_r\|_2\le\varrho$, and the genome lies in the box
$\Theta=[-B,B]^{Rp}\times[-B_U,B_U]^{pq}$.
\end{assumption}

All proofs are given in \Cref{app:proofs}.

\subsection{Positivity and existence}

\begin{theorem}[Positivity and metric equivalence]\label{thm:psd}
For every $\theta\in\Theta$ and $x\in\R^p$, $M_\theta(x)$ is symmetric and
\[
\epsilon\,I\ \preceq\ M_\theta(x)\ \preceq\ \Lambda\,I,\qquad
\Lambda:=e^{B}+pq\,B_U^2+\epsilon .
\]
Consequently, for every fixed $x$, $z\mapsto d_\theta(x,z)$ is the distance
induced by a norm, and
$\sqrt\epsilon\,\|x-z\|_2\le d_\theta(x,z)\le\sqrt\Lambda\,\|x-z\|_2$.
\end{theorem}

\begin{theorem}[Existence of an optimal metric]\label{thm:existence}
The objective $\mathcal J$ of \eqref{eq:objective} is continuous on
$\Theta$. Hence it attains its minimum: there is $\theta^\star\in\Theta$ with
$\mathcal J(\theta^\star)=\min_{\theta\in\Theta}\mathcal J(\theta)$.
\end{theorem}

The continuity of $\mathcal J$ is the reason why we use the stochastic
neighbour error instead of the $0$--$1$ error of the $k$-NN rule: the latter
is piecewise constant in $\theta$ and need not attain its infimum.

\subsection{Regularity of the metric field}

\begin{theorem}[Lipschitz metric field]\label{thm:lipschitz}
Under \Cref{ass:domain}, let $G:=4\varrho/\tau$. For all $x,x'\in\mathcal X$
and $\theta\in\Theta$,
\[
\|M_\theta(x)-M_\theta(x')\|_{\mathrm{op}}\ \le\ \mathrm{Lip}_M\,\|x-x'\|_2,
\qquad \mathrm{Lip}_M:=2BG\,e^{B}.
\]
\end{theorem}

Small temperatures $\tau$ (sharp gating) and large log-scales $B$ allow the
metric to change quickly; the smoothness penalty $\mathcal L_{\mathrm{sm}}$
controls the same quantity empirically on the training graph.

\subsection{Robustness to perturbations}

\begin{theorem}[Robustness of the neighbour structure]\label{thm:robust}
Under \Cref{ass:domain}, let $x,x+\delta\in\mathcal X$ and let $z$ range over
the training points. Then
\[
\big|d_\theta(x+\delta,z)-d_\theta(x,z)\big|\ \le\ \kappa\,\|\delta\|_2,
\qquad \kappa:=\sqrt\Lambda+\frac{\mathrm{Lip}_M\,\varrho}{\sqrt\epsilon}.
\]
Consequently:
(i) the $1$-NN prediction at $x$ is unchanged whenever
$2\kappa\|\delta\|_2<d^{\mathrm{miss}}_\theta(x)-d^{\mathrm{hit}}_\theta(x)$,
where $d^{\mathrm{hit}}$ ($d^{\mathrm{miss}}$) is the distance to the nearest
training point of the predicted class (of any other class);
(ii) the set of $k$ nearest neighbours of $x$, and hence the $k$-NN
prediction, is unchanged whenever
$2\kappa\|\delta\|_2<d_{(k+1)}(x)-d_{(k)}(x)$.
\end{theorem}

The theorem explains the role of the robustness term: $\mathcal L_{\mathrm{rob}}$
enlarges the relative hit/miss margins on the training points, while $\kappa$,
controlled by $B$, $B_U$, $\epsilon$ and the gating, bounds how fast distances
can move.

\subsection{Generalisation}

Let $\ell_\theta(x,y)=1-\sum_{j:y_j=y}p_j(x;\theta)$ with
$p_j(x;\theta)\propto\exp(-d_\theta^2(x,x_j))$ over the training points: this is
the probability of error of the stochastic neighbour rule, whose empirical
LOO version is $\mathcal L_{\mathrm{cls}}$. Write
$R(\theta)=\E\,\ell_\theta(X,Y)$ and $\hat R_m(\theta)$ for its average on an
i.i.d.\ sample of size $m$ independent of the training sample.

\begin{theorem}[Uniform generalisation]\label{thm:gen}
Under \Cref{ass:domain}, let $D=Rp+pq$, $\bar B=\max(B,B_U)$ and
$L:=4\varrho^2\,(e^{B}+2pq\,B_U)$. For every $\eta>0$ and $\delta\in(0,1)$,
with probability at least $1-\delta$, simultaneously for all
$\theta\in\Theta$,
\[
R(\theta)\ \le\ \hat R_m(\theta)+2\eta+\sqrt{\frac{D\ln\lceil\bar BL/\eta\rceil+\ln(2/\delta)}{2m}} .
\]
\end{theorem}

With $\eta=1/m$ the complexity term is of order
$\sqrt{D\log(m\bar BL)/m}$. It grows with the number of evolved parameters
$D=Rp+pq$, not with the number of training points, and it holds for any
$\hat\theta$ produced by any search procedure. This is the formal reason for
generating the local matrices from a small parametric family instead of
learning one free matrix per training point.

\subsection{Convergence of the evolutionary search}

\begin{theorem}[Convergence in value]\label{thm:ga}
Run the GA of \Cref{alg:eaml} without the stagnation stop. Let
$\mathcal J^{\mathrm{best}}_g$ be the best objective value found up to
generation $g$ and $\mathcal J^\star=\min_\Theta\mathcal J$. Then
$\mathcal J^{\mathrm{best}}_g$ is non-increasing and, for every $\eta>0$, there
is $\varepsilon_\eta>0$ with
$\Pb\{\mathcal J^{\mathrm{best}}_g>\mathcal J^\star+\eta\}\le(1-\varepsilon_\eta)^g$.
Hence $\mathcal J^{\mathrm{best}}_g\to\mathcal J^\star$ almost surely.
\end{theorem}

The result is asymptotic: $\varepsilon_\eta$ can be extremely small when $D$ is
large, so it guarantees that the search cannot stall at a non-optimal value
forever, not that a finite budget suffices. We measure the finite-budget
behaviour empirically in \Cref{sec:experiments}.

\begin{proposition}[Seeded dominance]\label{prop:seed}
The genome $\theta=0$ encodes the Euclidean metric up to the factor
$1+\epsilon$. Since it is in the initial population and replacement is
elitist, $\mathcal J(\hat\theta)\le\mathcal J(0)$: \eaml{} never does worse than
the Euclidean metric on its training objective.
\end{proposition}

\begin{proposition}[Cost]\label{prop:cost}
With $s\le n$ query rows in the fitness, one evaluation of $\mathcal J$ costs
$O(snp+snq+snR)$ time and $O(sn)$ memory, dominated by two matrix products.
Prediction costs $O(np+nq+Rp)$ per query.
\end{proposition}

\section{Experiments}\label{sec:experiments}

\subsection{Setup}

\paragraph{Datasets.} We use thirteen public datasets (\Cref{tab:datasets}).
Iris, Wine, Breast cancer and Digits ship with scikit-learn
\citep{pedregosa2011}; eight others come from the UCI repository \citep{uci}.
Fashion-MNIST \citep{xiao2017} is represented by a stratified subsample of
3000 test images. For this dataset the pixels are standardised and reduced to
40 principal components inside each training fold.

\paragraph{Methods.} All methods use the same $k$-NN rule, with $k$ chosen by
LOO accuracy on the training fold among $\{1,\dots,25\}$, so they differ only in
the metric. The baselines are:
Euclidean; Manhattan; Mahalanobis (regularised inverse covariance);
NCA \citep{goldberger2004} (scikit-learn, 100 iterations);
LMNN \citep{weinberger2009} and ITML \citep{davis2007} (metric-learn,
\citealp{devazelhes2020}, default settings, 3 target neighbours and 200
iterations for LMNN); and \textsc{GA-global}, which is \eaml{} with $R=1$ and
$q=0$, i.e.\ an evolved global diagonal metric in the spirit of
\citet{kelly1991,raymer2000}.
\eaml{} uses $R=4$, $q=2$, $B=3$, $B_U=1$ and $\epsilon=10^{-2}$. These values
were fixed before running the benchmark and are identical for all datasets.
LMNN could not be run on any Ecoli fold, because metric-learn rejects classes
with fewer training points than target neighbours, and ITML failed on one
Haberman fold. In these cases the method falls back to its initial
(Euclidean) metric, so on Ecoli LMNN coincides with the Euclidean baseline.

\paragraph{Protocol.} We use $5\times$ repeated stratified $5$-fold
cross-validation (25 splits per dataset, 325 in total). All preprocessing,
metric learning and the choice of $k$ take place inside the training fold
\citep{varma2006}. Besides accuracy and macro-F1, we measure robustness by
adding Gaussian noise $\mathcal N(0,\sigma^2I)$, $\sigma\in\{0.25,0.5\}$, to the
standardised test points. Across datasets we use the Friedman test, the
Nemenyi critical difference and Wilcoxon signed-rank tests with Holm correction
\citep{demsar2006,derrac2011}. Within a dataset we use the corrected resampled
$t$-test of \citet{nadeau2003}. Ablations are run on the first repetition (5
splits) of the eleven datasets other than Digits and Fashion-MNIST, for which
five additional \eaml{} runs per split were too costly.

\begin{table}[t]
\centering
\caption{Datasets: instances $n$, features $p$ (after removing constant ones; Fashion-MNIST is reduced by PCA inside each fold), classes $C$, and imbalance ratio.}\label{tab:datasets}
\small
\begin{tabular}{lrrrr}
\toprule
Dataset & $n$ & $p$ & $C$ & Imb.\ ratio\\
\midrule
Fashion-MNIST & 3000 & 783$\to$40 & 10 & 1.0 \\
Digits & 1797 & 61 & 10 & 1.1 \\
Iris & 150 & 4 & 3 & 1.0 \\
Wine & 178 & 13 & 3 & 1.5 \\
Breast cancer & 569 & 30 & 2 & 1.7 \\
Sonar & 208 & 60 & 2 & 1.1 \\
Ionosphere & 351 & 33 & 2 & 1.8 \\
Pima & 768 & 8 & 2 & 1.9 \\
Glass & 214 & 9 & 6 & 8.4 \\
Seeds & 210 & 7 & 3 & 1.0 \\
Banknote & 1372 & 4 & 2 & 1.2 \\
Haberman & 306 & 3 & 2 & 2.8 \\
Ecoli & 336 & 7 & 8 & 71.5 \\
 
\bottomrule
\end{tabular}
\end{table}

\subsection{Controlled study: a region-dependent relevant feature}\label{sec:synthetic}

We draw $x\in[-1,1]^{3+m}$ uniformly. The coordinate $x_0$ selects a region,
and the label is $y=\ind[\sin(2\pi x_1)>0]$ if $x_0<0$ and
$y=\ind[\sin(2\pi x_2)>0]$ otherwise. The other $m\in\{0,4,8\}$ coordinates are
nuisance features. A global metric has to keep both $x_1$ and $x_2$; a local
metric can drop the irrelevant one in each region. We train on 800 points,
test on 3000, and repeat 10 times.

\begin{table}[t]
\centering
\caption{Synthetic benchmark: test accuracy (\%, mean of 10 replications) for
$m$ nuisance features. $p$: Wilcoxon signed-rank test between \eaml{} and
\textsc{GA-global} (two-sided; the better method is in bold). Last columns:
mean local weights $e^{h_1},e^{h_2}$ learned by \eaml{} for $x_1$/$x_2$ in the
left ($x_0<0$) and right regions.}\label{tab:synthetic}
\small
\setlength{\tabcolsep}{4pt}
\begin{tabular}{ccccccccc}
\toprule
$m$ & Euclidean & NCA & LMNN & GA-global & \eaml{} & $p$ & left $x_1$/$x_2$ & right $x_1$/$x_2$\\
\midrule
0 & 90.95 & 89.91 & 90.90 & 91.40 & \textbf{93.22} & 0.002 & 13.95/5.95 & 5.20/14.99 \\
4 & 64.91 & 86.51 & 66.14 & \textbf{91.46} & 88.23 & 0.002 & 11.49/12.42 & 10.14/13.32 \\
8 & 58.47 & 75.34 & 58.94 & \textbf{90.43} & 83.12 & 0.002 & 10.17/10.88 & 9.70/11.04 \\
 
\bottomrule
\end{tabular}
\end{table}

\begin{figure}[t]
\centering
\includegraphics[width=\linewidth]{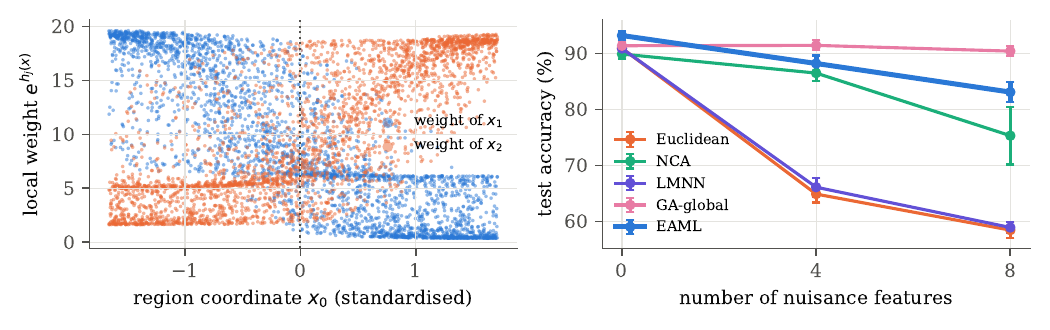}
\caption{Synthetic benchmark. Left: local weights of $x_1$ and $x_2$ learned by
\eaml{} on the test points of one replication without nuisance features,
plotted against the region coordinate $x_0$. The weights swap between the
regions. Right: accuracy versus number of nuisance features (mean $\pm$ s.d.).}
\label{fig:synthetic}
\end{figure}

Without nuisance features (\Cref{tab:synthetic}, \Cref{fig:synthetic}),
\eaml{} learns what the problem requires. On the left it gives $x_1$ a weight
about $2.3$ times that of $x_2$, and on the right the ratio is reversed. It is
the most accurate method (93.2\% against 91.4\% for \textsc{GA-global},
$p=0.002$). With nuisance features the order changes: \textsc{GA-global} stays
above 90\%, whereas \eaml{} falls to 88.2\% and 83.1\% (both $p=0.002$ in favour
of \textsc{GA-global}). The learned weights show why. With 4 or 8 nuisance
features, the weights of $x_1$ and $x_2$ no longer swap between the regions.
The $k$-means anchors are computed in the full noisy space, so they do not
align with the region boundary $x_0=0$. The genome also grows with the
dimension ($D=Rp+pq$) while the evaluation budget does not. Global metric
learners that ignore nuisance features (Euclidean, LMNN) collapse to about 59\%,
and NCA degrades to 75\%.

\subsection{Benchmark results}

\begin{table}[t]
\centering
\caption{Test accuracy (\%, mean\,$\pm$\,s.d.\ over 25 nested splits). Bold: best per row.
$^{\uparrow}$/$^{\downarrow}$: \eaml{} significantly better/worse than Euclidean (corrected resampled $t$-test, $\alpha=0.05$).
Mean rank: average Friedman rank (lower is better).}\label{tab:main}
\scriptsize
\setlength{\tabcolsep}{2.2pt}
\begin{tabular}{lcccccccc}
\toprule
Dataset & Euclidean & Manhattan & Mahalanobis & NCA & LMNN & ITML & GA-global & \eaml{}\\
\midrule
Fashion-MNIST & 80.2\,$\pm$\,1.3 & 80.3\,$\pm$\,1.3 & 80.2\,$\pm$\,1.3 & 76.6\,$\pm$\,1.7 & \textbf{80.8\,$\pm$\,1.0} & 80.5\,$\pm$\,1.1 & 79.8\,$\pm$\,1.1 & 79.9\,$\pm$\,1.4 \\
Digits & 97.6\,$\pm$\,0.5 & \textbf{98.0\,$\pm$\,0.6} & 95.6\,$\pm$\,1.2 & 97.8\,$\pm$\,0.6 & 98.0\,$\pm$\,0.7 & 98.0\,$\pm$\,0.8 & 97.7\,$\pm$\,0.9 & 97.5\,$\pm$\,0.7 \\
Iris & 96.3\,$\pm$\,3.5 & 94.5\,$\pm$\,4.1 & 90.3\,$\pm$\,4.5 & 95.9\,$\pm$\,3.9 & 96.3\,$\pm$\,3.5 & \textbf{96.9\,$\pm$\,2.5} & 94.5\,$\pm$\,3.6 & 94.5\,$\pm$\,4.4 \\
Wine & 96.3\,$\pm$\,2.5 & 97.2\,$\pm$\,2.3 & 93.5\,$\pm$\,3.4 & 96.6\,$\pm$\,2.9 & 96.4\,$\pm$\,3.3 & \textbf{98.3\,$\pm$\,2.3} & 95.7\,$\pm$\,2.8 & 97.4\,$\pm$\,2.9 \\
Breast cancer & 96.6\,$\pm$\,1.5 & \textbf{96.7\,$\pm$\,1.8} & 84.0\,$\pm$\,2.7 & 95.3\,$\pm$\,1.4 & \textbf{96.7\,$\pm$\,1.5} & 95.5\,$\pm$\,1.7 & 95.1\,$\pm$\,1.6 & 96.3\,$\pm$\,1.6 \\
Sonar & 84.9\,$\pm$\,5.1 & 85.2\,$\pm$\,5.2 & 76.5\,$\pm$\,5.6 & 86.7\,$\pm$\,3.9 & \textbf{87.3\,$\pm$\,4.7} & 78.3\,$\pm$\,4.8 & 85.0\,$\pm$\,6.4 & 84.6\,$\pm$\,6.2 \\
Ionosphere & 85.7\,$\pm$\,3.7 & \textbf{90.6\,$\pm$\,3.1} & 85.5\,$\pm$\,3.8 & 89.2\,$\pm$\,3.1 & 87.3\,$\pm$\,3.5 & 88.1\,$\pm$\,2.7 & 90.0\,$\pm$\,3.1 & 90.3\,$\pm$\,2.8$^{\uparrow}$ \\
Pima & 75.0\,$\pm$\,3.0 & 75.0\,$\pm$\,2.8 & 73.6\,$\pm$\,3.5 & 69.0\,$\pm$\,3.9 & 74.9\,$\pm$\,3.5 & 75.7\,$\pm$\,3.7 & \textbf{76.1\,$\pm$\,3.3} & 73.6\,$\pm$\,3.5 \\
Glass & 69.8\,$\pm$\,5.6 & 71.6\,$\pm$\,6.0 & 66.0\,$\pm$\,5.5 & 71.1\,$\pm$\,5.3 & 69.2\,$\pm$\,6.1 & 61.0\,$\pm$\,5.9 & \textbf{77.4\,$\pm$\,5.3} & 76.8\,$\pm$\,5.4$^{\uparrow}$ \\
Seeds & 92.5\,$\pm$\,3.7 & 92.3\,$\pm$\,3.2 & 94.3\,$\pm$\,3.3 & 93.2\,$\pm$\,3.7 & 92.5\,$\pm$\,3.8 & \textbf{95.8\,$\pm$\,2.9} & 93.0\,$\pm$\,2.9 & 91.7\,$\pm$\,3.5 \\
Banknote & 99.8\,$\pm$\,0.3 & 99.9\,$\pm$\,0.2 & \textbf{100.0\,$\pm$\,0.0} & \textbf{100.0\,$\pm$\,0.0} & 99.8\,$\pm$\,0.3 & \textbf{100.0\,$\pm$\,0.0} & 99.8\,$\pm$\,0.3 & 99.9\,$\pm$\,0.3 \\
Haberman & 74.8\,$\pm$\,3.7 & 74.4\,$\pm$\,3.6 & \textbf{75.0\,$\pm$\,4.1} & 71.0\,$\pm$\,6.3 & 74.8\,$\pm$\,3.5 & 74.4\,$\pm$\,3.4 & 74.6\,$\pm$\,4.3 & 73.2\,$\pm$\,3.8 \\
Ecoli & \textbf{86.3\,$\pm$\,3.6} & 85.2\,$\pm$\,3.2 & 85.9\,$\pm$\,3.5 & 74.0\,$\pm$\,14.4 & \textbf{86.3\,$\pm$\,3.6} & 85.9\,$\pm$\,3.5 & 84.8\,$\pm$\,4.2 & 84.6\,$\pm$\,3.4 \\
\midrule
Mean acc. & 87.37 & 87.76 & 84.65 & 85.88 & 87.72 & 86.80 & 87.98 & 87.72 \\
Mean rank & 4.65 & 3.65 & 5.69 & 5.00 & 3.54 & \textbf{3.46} & 4.77 & 5.23 \\
 
\bottomrule
\end{tabular}
\end{table}

\begin{table}[t]
\centering
\caption{Pairwise comparison of \eaml{} with each baseline over the 13 datasets:
wins/ties/losses, Wilcoxon $p$-value, Holm-adjusted $p$-value, and number of
datasets where \eaml{} is significantly better/worse (corrected resampled $t$-test).}\label{tab:stats}
\small
\begin{tabular}{lcccc}
\toprule
Baseline & W/T/L & $p$ & $p_{\text{Holm}}$ & sig.\ better/worse\\
\midrule
Euclidean & 4/0/9 & 0.414 & 1.000 & 2/0 \\
Manhattan & 3/1/9 & 0.062 & 0.438 & 0/0 \\
Mahalanobis & 7/0/6 & 0.110 & 0.659 & 4/0 \\
NCA & 8/0/5 & 0.146 & 0.732 & 2/0 \\
LMNN & 4/0/9 & 0.376 & 1.000 & 2/0 \\
ITML & 4/0/9 & 0.635 & 1.000 & 1/1 \\
GA-global & 5/1/7 & 0.395 & 1.000 & 0/0 \\
 
\bottomrule
\end{tabular}
\end{table}

\Cref{tab:main,tab:stats} summarise the benchmark. Overall, no method stands
out: the Friedman test does not reject equal performance ($\chi^2_F=11.0$,
$p=0.14$). ITML (3.46), LMNN (3.54) and Manhattan (3.65) have the best mean
ranks. \textsc{GA-global} has the highest mean accuracy (87.98\%) but a middle
rank (4.77). \eaml{} ranks 5.23, with a mean accuracy of 87.72\%. The
differences between \eaml{} and every baseline are not significant across
datasets (all Holm-adjusted $p\ge0.44$).

The local model helps on two datasets. On Glass, \eaml{} improves on the
Euclidean metric by $7.0$ points (76.8\% vs.\ 69.8\%), and on Ionosphere by
$4.6$ points; both differences are significant. On these datasets the global
evolved metric does about as well (77.4\% and 90.0\%), so the gain mainly comes
from evolving feature weights rather than from locality. Elsewhere \eaml{}
loses up to $1.7$ points to the Euclidean metric (Iris, Ecoli, Haberman,
Pima), although no loss is significant.

\begin{table}[t]
\centering
\caption{Robustness to Gaussian noise added to the standardised test points (13 datasets).}\label{tab:robust}
\scriptsize
\setlength{\tabcolsep}{2.5pt}
\begin{tabular}{lcccccccc}
\toprule
 & Euclidean & Manhattan & Mahalanobis & NCA & LMNN & ITML & GA-global & \eaml{}\\
\midrule
Mean acc., $\sigma=0$ & 87.37 & 87.76 & 84.65 & 85.88 & 87.72 & 86.80 & 87.98 & 87.72 \\
Mean rank, $\sigma=0$ & 4.65 & 3.65 & 5.69 & 5.00 & 3.54 & \textbf{3.46} & 4.77 & 5.23 \\
Mean acc., $\sigma=0.25$ & 86.08 & 86.54 & 77.71 & 83.46 & 86.22 & 84.24 & 85.80 & 86.02 \\
Mean rank, $\sigma=0.25$ & 3.77 & \textbf{2.69} & 6.54 & 5.96 & 2.92 & 5.04 & 4.54 & 4.54 \\
Mean acc., $\sigma=0.5$ & 83.57 & 83.49 & 70.31 & 79.39 & 83.62 & 78.44 & 81.57 & 81.93 \\
Mean rank, $\sigma=0.5$ & \textbf{2.19} & 2.69 & 6.77 & 6.00 & 2.42 & 6.27 & 4.88 & 4.77 \\
Mean drop at $\sigma=0.5$ & 3.80 & 4.27 & 14.34 & 6.49 & 4.10 & 8.36 & 6.41 & 5.79 \\
 
\bottomrule
\end{tabular}
\end{table}

\begin{figure}[t]
\centering
\includegraphics[width=0.55\linewidth]{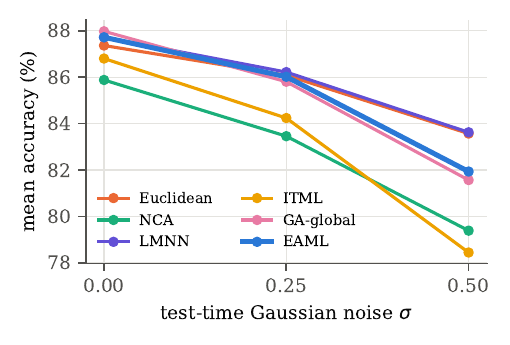}
\caption{Mean accuracy over the 13 datasets as a function of the test-time noise level.}
\label{fig:robust}
\end{figure}

\paragraph{Robustness.} \Cref{tab:robust,fig:robust} show that the robustness
term does not translate into better robustness to test-time noise. At
$\sigma=0.5$, \eaml{} loses $5.8$ points on average, against $3.8$ for the
Euclidean metric and $4.1$ for LMNN. It nevertheless degrades less than
\textsc{GA-global} ($6.4$), NCA ($6.5$), ITML ($8.4$) and Mahalanobis ($14.3$).
The main reason is that the learned metrics are sharp: the stochastic-neighbour
term rewards metrics under which the nearest same-class point dominates. As a
result, \eaml{} and \textsc{GA-global} select $k=1$ in most folds (median $k=1$, as
for NCA, against 4 to 6 for the other metrics, \Cref{tab:other}), and the 1-NN rule is
the most sensitive to noise. This is consistent with \Cref{thm:robust}: the
guarantee depends on the hit/miss margin \emph{relative} to $\kappa$, a ratio
that a mere rescaling of $M_\theta$ leaves unchanged, so sharper metrics do not
buy robustness by themselves.

\begin{table}[t]
\centering
\caption{Macro-F1, wall-clock training time per fold (one CPU core) and median selected $k$.}\label{tab:other}
\scriptsize
\setlength{\tabcolsep}{2.5pt}
\begin{tabular}{lcccccccc}
\toprule
 & Euclidean & Manhattan & Mahalanobis & NCA & LMNN & ITML & GA-global & \eaml{}\\
\midrule
Macro-F1 (mean) & 84.04 & 84.25 & 80.86 & 82.41 & 84.40 & 82.86 & 84.91 & 84.28 \\
Macro-F1 (rank) & 4.69 & 3.92 & 6.00 & 5.00 & 3.54 & \textbf{3.46} & 4.46 & 4.92 \\
Fit time, median over datasets (s) & 0.00 & 0.00 & 0.00 & 0.09 & 0.51 & 0.25 & 3.62 & 3.93 \\
Fit time, Fashion-MNIST (s) & 0.2 & 0.1 & 0.1 & 22.7 & 9.1 & 29.7 & 162.2 & 171.1 \\
Median selected $k$ & 6 & 5 & 4 & 1 & 6 & 6 & 1 & 1 \\
 
\bottomrule
\end{tabular}
\end{table}

\paragraph{Cost.} The median training time of \eaml{} is $3.9$\,s per fold,
against less than $0.6$\,s for NCA, LMNN and ITML. On the largest problem
(Fashion-MNIST, 2400 training points, 40 dimensions) it takes $171$\,s, against
9--30\,s for the gradient-based learners.

\subsection{Search behaviour}

\begin{figure}[t]
\centering
\includegraphics[width=\linewidth]{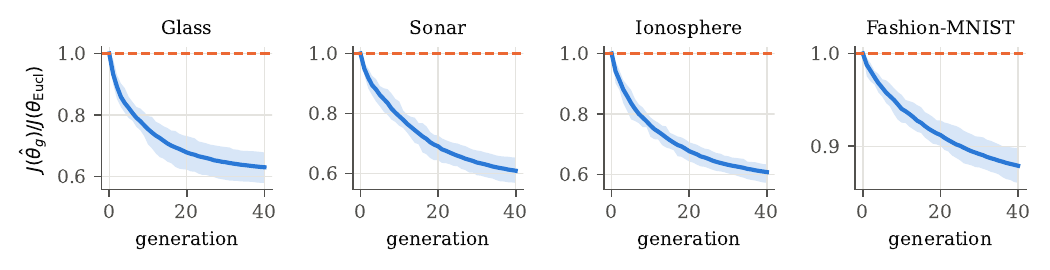}
\caption{Best objective value relative to the Euclidean genome, per generation (mean and 10--90\% band over 25 folds). The dashed line is the seeded starting point (\Cref{prop:seed}).}
\label{fig:convergence}
\end{figure}

The GA reduces the training objective to $0.58$ times its Euclidean value on
average, from $0.08$ on Banknote to $0.89$ on Haberman. However, every run used
all 40 generations without triggering the stagnation stop, and the curves in
\Cref{fig:convergence} are still decreasing at the end. The search budget is
therefore binding. This is consistent with the asymptotic nature of
\Cref{thm:ga}: almost-sure convergence in value says nothing about the speed
at which a 366-dimensional genome (Digits) is optimised with 1230 evaluations.

\subsection{Ablation study}

\begin{table}[t]
\centering
\caption{Ablations (first repetition, 11 datasets): mean accuracy (\%), mean rank among the six variants, mean accuracy at $\sigma=0.5$, objective relative to the Euclidean genome, final $\mathcal L_{\mathrm{cls}}$ and $\mathcal L_{\mathrm{rob}}$, median fit time (s).}\label{tab:ablation}
\small
\setlength{\tabcolsep}{4pt}
\begin{tabular}{lccccccc}
\toprule
Variant & Acc. & Rank & Acc.\ $\sigma{=}0.5$ & $\mathcal J/\mathcal J_{\mathrm{Eucl}}$ & $\mathcal L_{\mathrm{cls}}$ & $\mathcal L_{\mathrm{rob}}$ & Time\\
\midrule
EAML & 87.95 & 2.88 & 81.70 & 0.577 & 0.114 & 0.293 & 3.8 \\
EAML (no robustness) & 87.25 & 4.14 & 81.16 & 0.477 & 0.113 & 0.332 & 3.3 \\
EAML (no smoothness) & 88.33 & 2.50 & 81.48 & 0.543 & 0.117 & 0.289 & 2.9 \\
EAML (no low-rank, q=0) & 87.97 & 3.95 & 80.34 & 0.565 & 0.119 & 0.299 & 3.2 \\
EAML (random search) & 87.31 & 3.23 & 80.69 & 0.684 & 0.138 & 0.331 & 1.9 \\
EAML (NCA term only) & 87.55 & 3.95 & 81.53 & 0.442 & 0.113 & 0.331 & 3.1 \\
 
\bottomrule
\end{tabular}
\end{table}

\Cref{tab:ablation} isolates the design choices. The numbers in the $\mathcal J$
column are not comparable across rows when the objective itself differs.
\begin{itemize}
\item \textbf{Robustness term.} Removing it lowers accuracy ($87.25\%$ vs.\
  $87.95\%$) and the noisy accuracy ($81.16\%$ vs.\ $81.70\%$), and gives the
  worst rank. This is the only term whose removal clearly hurts.
\item \textbf{Smoothness term.} Removing it slightly \emph{improves} accuracy
  ($88.33\%$, best rank), so with our default weight the smoothness prior is too
  strong for these datasets.
\item \textbf{Low-rank term.} With $q=0$ the clean accuracy is unchanged but the
  noisy accuracy drops by $1.4$ points.
\item \textbf{Search.} Random search with the same budget reaches a worse
  objective ($0.68$ vs.\ $0.58$ of the Euclidean value) and a lower accuracy
  ($87.31\%$). The genetic operators do matter.
\item \textbf{NCA term only.} Optimising only $\mathcal L_{\mathrm{cls}}$ lowers
  accuracy by $0.4$ points.
\end{itemize}

\begin{figure}[t]
\centering
\includegraphics[width=0.75\linewidth]{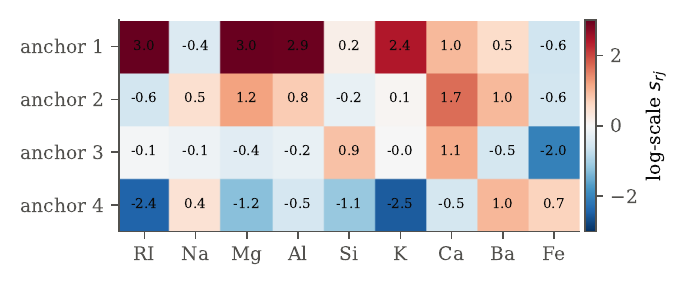}
\caption{Log-scales $s_r$ learned at the four anchors on Glass (first fold). Positive values increase the weight of a feature near that anchor.}
\label{fig:anchors}
\end{figure}

\paragraph{Learned geometry.} \Cref{fig:anchors} shows the anchor log-scales
learned on Glass, the dataset where \eaml{} gains most. The anchors disagree
substantially. For example, the refractive index (RI) and potassium (K) are
strongly up-weighted near anchor 1 and down-weighted near anchor 4. This
confirms that the model does use its local degrees of freedom when the data
allow it.

\FloatBarrier
\section{Discussion and limitations}\label{sec:discussion}

\paragraph{When do local metrics pay off?} Locality helps when (i) the
relevant directions genuinely change across the space and (ii) the gating can
see the regions. The controlled study meets both conditions without nuisance
features, and \eaml{} then outperforms every global metric. When nuisance
dimensions blur the unsupervised gating, or when the data are close to
homogeneous, the additional degrees of freedom cost more search effort and
variance than they bring. A global evolved metric is then preferable. On the
thirteen benchmarks, no metric learner, global or local, significantly beats
the others, which is a useful reminder that careful nested evaluation of
metric learning often yields small differences.

\paragraph{Limitations.} (i) The search budget is binding
(\Cref{fig:convergence}). Our convergence theorem is asymptotic and does not
bound the number of generations. (ii) The anchors are unsupervised; supervised
or evolved anchors, or a gating on the learned metric itself, are natural
fixes for the misalignment observed with nuisance features. (iii) The
stochastic-neighbour term produces sharp metrics and $k=1$, which hurts
robustness. A temperature or margin constraint on the scale of $M$ would
counter this. (iv) The generalisation bound concerns the stochastic neighbour
risk on an independent sample, not the LOO-selected majority vote. (v) The
objective is a fixed scalarisation; a Pareto approach (e.g.\ NSGA-II,
\citealp{deb2002}) would expose the accuracy--robustness trade-off directly.
(vi) Training is one to two orders of magnitude slower than gradient-based
metric learners.

\paragraph{Future work.} The results point to three concrete steps: a larger
or adaptive evaluation budget with warm starts from \textsc{GA-global};
supervised placement of anchors; and multi-objective evolution of local
metrics, with the theory above applied to each Pareto-optimal solution.

\section{Conclusion}

We introduced \eaml{}, an evolutionary framework that learns a field of local
Mahalanobis metrics from a small set of generating parameters, under a
continuous multi-objective criterion. The framework comes with guarantees on
positivity, existence, regularity, robustness, generalisation and convergence
in value of the search. Experiments show that local metrics recover
region-dependent geometry when it exists and the regions are identifiable. On
standard benchmarks their benefit is confined to a few datasets and does not
survive nuisance features or test-time noise, where a global evolved metric is
more reliable. These positive and negative findings delimit the regime in
which evolutionary local metric learning is worth its cost.

\paragraph{Code and data availability.} The Python implementation of \eaml{},
the baselines and the scripts that reproduce every table and figure of this
paper are available from the corresponding author upon request. All datasets
used in this study are publicly available (\Cref{tab:datasets}).

\FloatBarrier
\def\bibfont{\raggedright}
\bibliographystyle{plainnat}
\bibliography{references}

\appendix
\section{Proofs}\label{app:proofs}

\begin{proof}[Proof of \Cref{thm:psd}]
$D_\theta(x)$ is diagonal with entries $e^{h_j(x)}>0$, and $UU^\top\succeq0$, so
$M_\theta(x)\succeq\epsilon I$. Since $h_\theta(x)$ is a convex combination of
vectors in $[-B,B]^p$, every entry of $D_\theta(x)$ is at most $e^B$. Moreover
$\|UU^\top\|_{\mathrm{op}}=\|U\|_{\mathrm{op}}^2\le\|U\|_F^2\le pqB_U^2$. Hence
$M_\theta(x)\preceq\Lambda I$. For fixed $x$, $z\mapsto\|M_\theta(x)^{1/2}(x-z)\|_2$
is a norm of $x-z$, and the two-sided bound follows from the eigenvalue bounds.
\end{proof}

\begin{proof}[Proof of \Cref{thm:existence}]
For fixed $(i,j)$, $d^2_\theta(x_i,x_j)$ is a finite sum of products of
exponentials and polynomials in $\theta$, hence continuous. $\mathcal L_{\mathrm{cls}}$
is a finite combination of $\log\sum\exp$ of these quantities, hence continuous.
Each $d^{\mathrm h}_i$ and $d^{\mathrm m}_i$ is the square root of a minimum of
finitely many continuous functions, hence continuous. (If a class has a single
training point, $d^{\mathrm h}_i$ is set to the constant $0$.) Since the
implemented ratio $\rho_i$ has denominator $d^{\mathrm m}_i+d^{\mathrm h}_i+10^{-12}>0$,
$\mathcal L_{\mathrm{rob}}$ is continuous. $\mathcal L_{\mathrm{sm}}$ and $\mathcal C$ are
polynomials in $\theta$ (the gating does not depend on $\theta$). Thus
$\mathcal J$ is continuous on the compact box $\Theta$, and the Weierstrass
theorem gives a minimiser.
\end{proof}

\begin{proof}[Proof of \Cref{thm:lipschitz}]
Write $a_r(x)=-\|x-c_r\|^2/\tau$ and $g_r=\nabla a_r=-2(x-c_r)/\tau$, so that
$\|g_r\|\le2(\|x\|+\|c_r\|)/\tau\le 4\varrho/\tau=G$. The softmax gives
$\nabla\pi_r=\pi_r(g_r-\bar g)$ with $\bar g=\sum_s\pi_sg_s$, hence
$\|\nabla\pi_r\|\le2G\pi_r$. Therefore
$\|\nabla h_j\|\le\sum_r|s_{rj}|\,\|\nabla\pi_r\|\le 2BG$, and by the mean value
theorem on the convex set $\mathcal X$,
$|h_j(x)-h_j(x')|\le2BG\|x-x'\|$. Since $|h_j|\le B$,
$|e^{h_j(x)}-e^{h_j(x')}|\le e^B|h_j(x)-h_j(x')|$. The terms $UU^\top$ and
$\epsilon I$ do not depend on $x$, and the operator norm of a diagonal matrix is
its largest absolute entry, which gives the claim.
\end{proof}

\begin{proof}[Proof of \Cref{thm:robust}]
Let $x'=x+\delta$, $a=(x'-z)^\top M_\theta(x')(x'-z)$ and
$b=(x'-z)^\top M_\theta(x)(x'-z)$. By the triangle inequality,
\[
|d_\theta(x',z)-d_\theta(x,z)|\le|\sqrt a-\sqrt b|+\big|\|x'-z\|_{M_\theta(x)}-\|x-z\|_{M_\theta(x)}\big| .
\]
The second term is at most $\|\delta\|_{M_\theta(x)}\le\sqrt\Lambda\|\delta\|$
(\Cref{thm:psd}). For the first, $a,b\ge\epsilon\|x'-z\|^2$, so
$|\sqrt a-\sqrt b|=|a-b|/(\sqrt a+\sqrt b)\le|a-b|/(2\sqrt\epsilon\|x'-z\|)$. By
\Cref{thm:lipschitz}, $|a-b|\le\mathrm{Lip}_M\|\delta\|\,\|x'-z\|^2$, so the first
term is at most $\mathrm{Lip}_M\|\delta\|\,\|x'-z\|/(2\sqrt\epsilon)\le\mathrm{Lip}_M\varrho\|\delta\|/\sqrt\epsilon$,
because $\|x'-z\|\le2\varrho$. This proves the bound with $\kappa$. Every
distance from the query to a training point therefore moves by at most
$\kappa\|\delta\|$. If two such distances differ by more than $2\kappa\|\delta\|$,
their order is preserved. Applied to the nearest hit and miss, this gives (i);
applied to the $k$-th and $(k+1)$-th neighbours, it shows that the $k$-NN set,
and hence the vote, is unchanged, which gives (ii).
\end{proof}

\begin{proof}[Proof of \Cref{thm:gen}]
\emph{Lipschitz property in $\theta$.} Fix $(x,y)$ and let $u_j=-d^2_\theta(x,x_j)$.
Writing $P_S=\sum_{j\in S}p_j$ with $S=\{j:y_j=y\}$, we have
$\partial P_S/\partial u_l=p_l(\ind[l\in S]-P_S)$, so
$\|\nabla_u\ell\|_1\le\sum_lp_l=1$ and
$|\ell(u)-\ell(u')|\le\|u-u'\|_\infty$. Next, with $\Delta=x-x_j$ and
$\|\Delta\|\le2\varrho$,
$\partial d^2/\partial s_{rf}=\pi_r e^{h_f}\Delta_f^2$ and
$\partial d^2/\partial U_{fl}=2\Delta_f(U^\top\Delta)_l$. Summing absolute values,
$\sum_{r,f}\pi_re^{h_f}\Delta_f^2\le e^B\|\Delta\|^2\le4\varrho^2e^B$ and
$\sum_{f,l}2|\Delta_f||(U^\top\Delta)_l|\le2qB_U\|\Delta\|_1^2\le2pqB_U\|\Delta\|^2\le8pqB_U\varrho^2$.
Hence $\|\nabla_\theta d^2\|_1\le L$, and for $\theta,\theta'\in\Theta$ (a convex
set), $|d^2_\theta-d^2_{\theta'}|\le L\|\theta-\theta'\|_\infty$. Combining both
steps, $|\ell_\theta(x,y)-\ell_{\theta'}(x,y)|\le L\|\theta-\theta'\|_\infty$ for all
$(x,y)$, so the same holds for $R$ and $\hat R_m$.

\emph{Covering and union bound.} A grid with $\lceil\bar BL/\eta\rceil$ points per
coordinate covers $\Theta$ within $\ell_\infty$-distance $\eta/L$, so it has
$N\le\lceil\bar BL/\eta\rceil^D$ points. Since $\ell\in[0,1]$, Hoeffding's
inequality \citep{hoeffding1963} and a union bound give, with probability at
least $1-\delta$, $|R(\theta')-\hat R_m(\theta')|\le\sqrt{\ln(2N/\delta)/(2m)}$ for
all grid points $\theta'$. For any $\theta$, choosing the closest grid point
changes $R$ and $\hat R_m$ by at most $\eta$ each, which yields the bound.
\end{proof}

\begin{proof}[Proof of \Cref{thm:ga}]
\emph{Monotonicity.} The best parents enter the truncated merged population, so
the best value never increases.

\emph{Hitting probability.} Fix $\eta>0$ and let
$A_\eta=\{\theta\in\Theta:\mathcal J(\theta)<\mathcal J^\star+\eta\}$. By
continuity (\Cref{thm:existence}), $A_\eta$ is relatively open in $\Theta$ and
contains a minimiser, so $\mathrm{Leb}(A_\eta)>0$. Consider any offspring, with
pre-mutation genome $\tilde\theta\in\Theta$. All $D$ genes are mutated with
probability at least $0.1^{D-1}$: one gene is forced, and the others mutate
independently with probability $0.1$. Conditionally on this, the offspring is the
coordinate-wise reflection into $\Theta$ of
$\tilde\theta+\sigma_gS\xi$, $\xi\sim\mathcal N(0,I_D)$, where $S$ holds the box widths
$w_j$ and $\sigma_g\in[\sigma_{\min},\sigma_0]$. Its density at any $t\in\Theta$ is
at least that of the unreflected Gaussian. Since $|t_j-\tilde\theta_j|\le w_j$,
this density is at least
$c=\prod_j(\sqrt{2\pi}\sigma_0w_j)^{-1}e^{-1/(2\sigma_{\min}^2)}>0$. Hence each
offspring lands in $A_\eta$ with probability at least
$\varepsilon_\eta=0.1^{D-1}c\,\mathrm{Leb}(A_\eta)>0$, whatever the past. With at
least one offspring per generation,
$\Pb\{\mathcal J^{\mathrm{best}}_g\ge\mathcal J^\star+\eta\}\le(1-\varepsilon_\eta)^g$.

\emph{Almost-sure convergence.} By Borel--Cantelli,
$\mathcal J^{\mathrm{best}}_g<\mathcal J^\star+\eta$ eventually, almost surely, for
each rational $\eta>0$. Intersecting over these countably many events and using
monotonicity gives $\mathcal J^{\mathrm{best}}_g\to\mathcal J^\star$ almost surely.
\end{proof}

\begin{proof}[Proof of \Cref{prop:seed,prop:cost}]
For $\theta=0$, $h_\theta\equiv0$, $D_\theta\equiv I$ and $U=0$, so
$M_\theta=(1+\epsilon)I$. The seed is evaluated at generation $0$, and by
monotonicity the returned genome satisfies $\mathcal J(\hat\theta)\le\mathcal J(0)$.
For the cost, the squared distances are
$d^2_{ij}=\sum_fE_{if}x_{if}^2-2\sum_fE_{if}x_{if}x_{jf}+\sum_fE_{if}x_{jf}^2+\dots$,
with $E=e^{\Pi S}$. This requires the products $(E\circ X_S)X^\top$ and
$E(X\circ X)^\top$ ($O(snp)$), the low-rank term ($O(snq)$), the gating
products ($O(sRp)$) and row-wise reductions ($O(sn)$).
\end{proof}

\section{Hyper-parameters}\label{app:hyper}

\begin{table}[h]
\centering
\caption{Default settings of \eaml{} (identical for all datasets).}
\small
\begin{tabular}{llll}
\toprule
Setting & Value & Setting & Value\\
\midrule
anchors $R$, rank $q$ & $4$, $2$ & population, generations & $30$, $40$ (patience 15)\\
bounds $B$, $B_U$ & $3$, $1$ & tournament size & $3$\\
$\epsilon$ & $10^{-2}$ & SBX probability, index & $0.9$, $15$\\
$\lambda_1,\lambda_2,\lambda_3$ & $0.5$, $0.1$, $0.01$ & mutation probability per gene & $0.1$\\
margin target $\bar\rho$ & $0.2$ & mutation step $\sigma_0\to\sigma_{\min}$ & $0.25\to0.02$\\
query rows $s$ & $\min(n,800)$ & elites & $2$\\
$k$ search range & $\{1,\dots,25\}$ & smoothness graph & Euclidean 5-NN\\
\bottomrule
\end{tabular}
\end{table}

\end{document}